\documentclass[letterpaper, 10 pt, conference]{ieeeconf}  

\IEEEoverridecommandlockouts                              

\usepackage{graphics} 
\usepackage{epsfig} 
\usepackage{mathptmx} 
\usepackage{times} 
\usepackage{amsmath} 
\usepackage{amssymb}  
\usepackage{cite}
\usepackage{booktabs}
\usepackage{stfloats}
\usepackage{caption}
\usepackage{lipsum}
\usepackage{wrapfig}
\usepackage{multirow}
\usepackage{graphicx}
\usepackage{makecell}
\usepackage[hyphens]{url}
\usepackage{xurl}
\usepackage{xcolor}
\usepackage{hyperref}
\hypersetup{
  breaklinks=true,
}
\usepackage{pgfplots}
\pgfplotsset{compat=1.17}
\newcommand{\method}{\texttt{VLA${^{\texttt 2}}$}}
\usepackage{circledsteps}
\usepackage{listings}
\title{\LARGE \bf
\method: Empowering \underline{V}ision-\underline{L}anguage-\underline{A}ction Models with an \underline{A}gentic Framework for Unseen Concept Manipulation
}

\author{Han Zhao\textsuperscript{\ddag\thanks{\ddag Equal Contribution},1,2}, 
Jiaxuan Zhang\textsuperscript{\ddag,2,3}, 
Wenxuan Song\textsuperscript{4}, 
Pengxiang Ding\textsuperscript{1,2}, 
Donglin Wang\textsuperscript{*,2\thanks{*Corresponding Author}},\\
\normalsize \textsuperscript{1}Zhejiang University, China
\normalsize \textsuperscript{2}MiLAB, Westlake University, China \\
\normalsize \textsuperscript{3}Southern University of Science and Technology, China \\
\normalsize \textsuperscript{4}Hong Kong University of Science and Technology (Guangzhou), China \\}

\begin{document}

\maketitle
\thispagestyle{empty}
\pagestyle{empty}

\begin{abstract}
Current vision-language-action (VLA) models, pre-trained on large-scale robotic data, exhibit strong multi-task capabilities and generalize well to variations in visual and language instructions for manipulation. However, their success rate drops significantly when faced with object concepts outside the training data, such as unseen object descriptions and textures in the dataset.
To address this, we propose a novel agentic framework, \method, which leverages OpenVLA as the execution backbone and effectively leverages external modules such as web retrieval and object detection to provide visual and textual knowledge about target objects to the VLA. This approach mitigates generalization failure when handling out-of-distribution objects.
Based on the LIBERO simulation environment, we introduced novel objects and object descriptions to construct a new evaluation benchmark with three difficulty levels to test the effectiveness of our method. Our framework successfully outperformed the current state-of-the-art models on our designed hard-level generalization benchmark. Compared to the standalone OpenVLA baseline, \method~achieves a 44.2\% improvement in the success rate in the hard-level benchmark and an average improvement of 20.2\% in all customized environments without any performance degradation on in-domain tasks. Project website: \hyperref[https://vla-2.github.io]{https://vla-2.github.io}.
\end{abstract}

\section{INTRODUCTION}
In recent years, foundation models have profoundly influenced the development of artificial intelligence research. This impact spans visual encoders~\cite{yang2025clip, caron2021dino, zhai2023siglip}, multi-modal large language models~\cite{liu2023llava, karamcheti2024prismatic, Zhao2025Cobra}, and agent systems~\cite{wang2023voyager, gur2024webagent, yao2022react}, among others. In the field of robotics, Vision-Language-Action (VLA) models~\cite{brohan2023rt2, kim2024openvla, black2024pi0, Ding2025, zhou2025chatvla, cheang2025gr3, nvidia2025gr00t} built upon vision-language models represent a prominent research paradigm. By fully integrating visual perception, language instruction understanding, and action execution into a unified model, VLA leverages large-scale robotic manipulation datasets for end-to-end training. This approach effectively harnesses the learning capacity of large-scale models and shows strong potential to serve as a foundational backbone for general-purpose robots performing manipulation tasks in open-world environments in the future.

However, although VLA models have acquired a certain degree of generalization ability, such as understanding some unseen language instructions and manipulating corresponding objects, they completely fail to comprehend instructions involving entirely unseen concepts (as demonstrated in OpenVLA failure cases~\cite{kim2024openvla}) and are unable to transfer previously learned manipulation experience to such scenarios. Some researchers have attempted to jointly train robotic manipulation data with web-scale multimodal data~\cite{zhou2025chatvla, brohan2023rt2}, aiming to preserve extensive conceptual knowledge during training and thereby enhance generalization in manipulation tasks. However, such a training paradigm not only demands substantial resources but also makes iterative model updates with emerging concepts impractical. As a result, it fails to fully address the problem.

\begin{figure}[t] 
  \centering
  \includegraphics[width=0.45\textwidth]{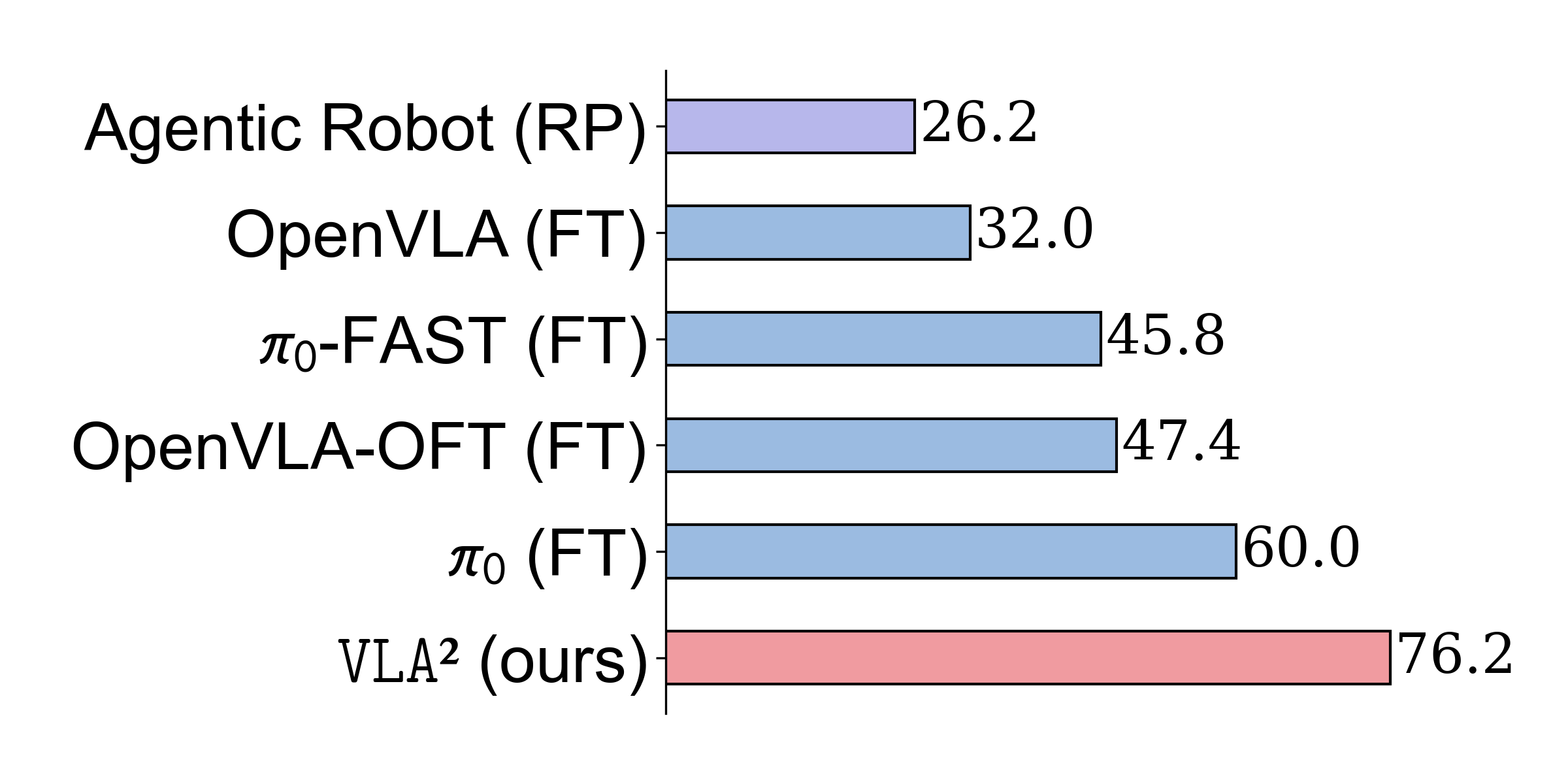}
  \caption{\textbf{Evaluation result on our custom Hard-level benchmark.} In evaluation involving unseen concepts (i.e., object textures and language descriptions outside the dataset), our proposed framework surpasses other state-of-the-art models finetuned on the original LIBERO dataset. In contrast, the reproduced Agentic Robot framework~\cite{Yang2025} using our model exhibits a significantly noticeable performance degradation in this task.}
  \label{fig:hard-compare}
\end{figure}

To this end, we proposed \textbf{V}ision-\textbf{L}anguage-\textbf{A}ction \textbf{A}gent (\method), a novel integrated system-level framework designed to increase the capabilities of VLA systems by supporting the invocation of diverse tools—including task planning, web search, object detection, and other functional modules—thereby extending the executive limits of the current VLA models.

Our main contributions are as follows:

\begin{itemize}
    \item We propose the \method~framework that integrates task planning, conversion of unseen concepts into known information via web and memory retrieval, VLA-based execution, and a verifier module to assess task completion.
    \item We fine-tune OpenVLA\cite{kim2024openvla} on the augmented LIBERO\cite{liu2023libero} dataset to enable the VLA to accept masked images as input conditions for improving generalization in object manipulation.
    \item Based on the LIBERO simulation environment, we designed object generalization tasks across three difficulty levels, ranging from simple color variations (Easy) and manipulation of generalized target objects (Medium) to generalization to objects with unseen concepts (Hard).
\end{itemize}

\section{RELATED WORKS}
\subsection{Vision-Language-Action Models}
VLA models~\cite{brohan2023rt2, kim2024openvla, black2024pi0, pertsch2025fast, Ding2025, zhou2025chatvla, cheang2025gr3, nvidia2025gr00t} belong to a type of foundation model that processes visual and other modal data as observations, and follows human natural language commands to execute the corresponding robotic tasks. Through pre-training on large-scale robotic manipulation datasets~\cite{khazatsky2025droid, embodimentcollaboration2025openx, agibotworldcontributors2025} and minimal fine-tuning through supervised fine-tuning~\cite {kim2025oft, li2025controlvla, song2025ceedvla} or reinforcement learning~\cite{Song2024germ, zhao2025more, lu2025, zhang2025reinbot, tan2025, chen2025conrft} on downstream tasks.

\begin{figure*}[t] 
  \centering
  \includegraphics[width=0.9\textwidth]{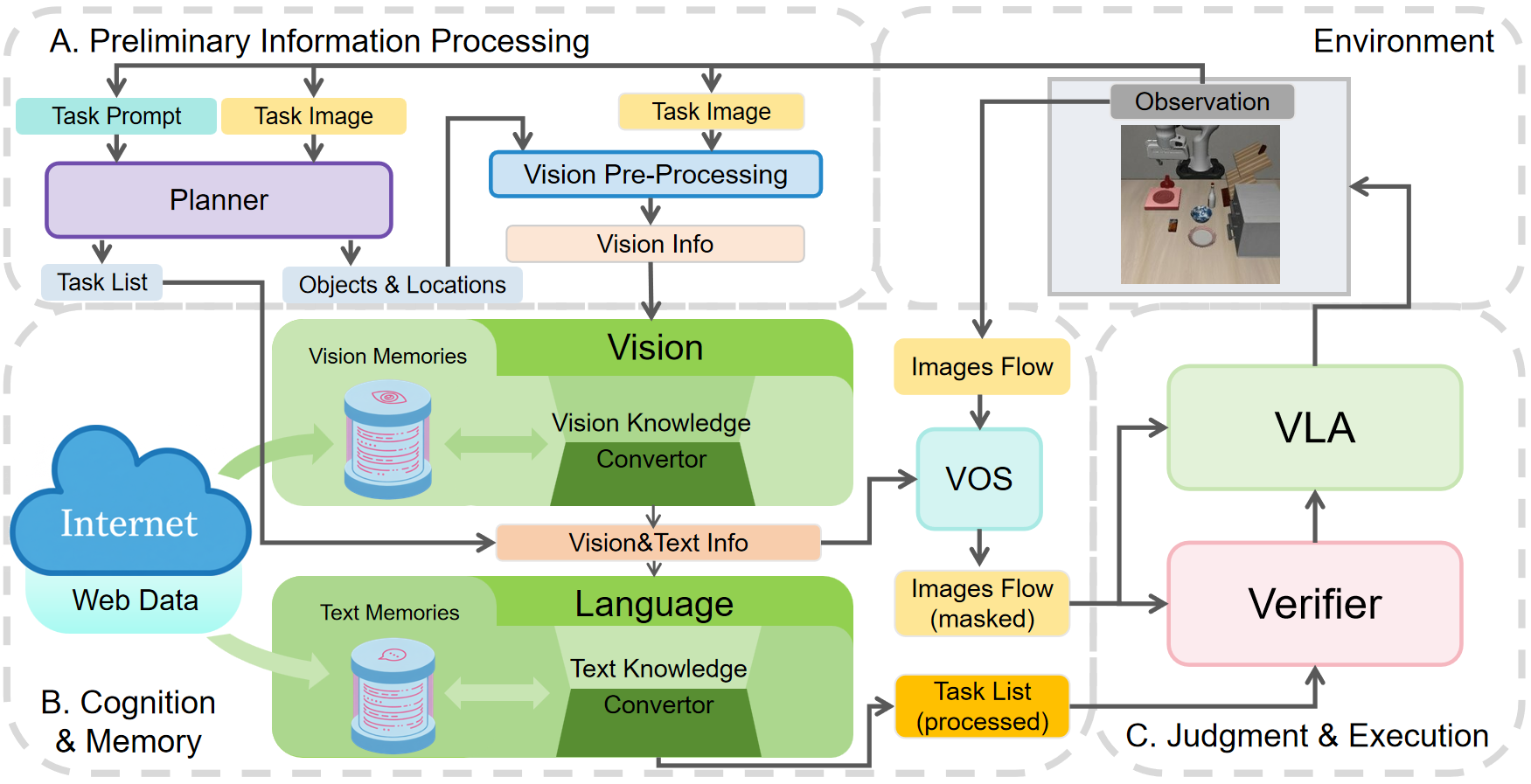} 
  \caption{\textbf{Framework overview}. The proposed framework comprises three components: A. preliminary processing, B. cognition and memory, and C. Judgment and execution. During task running, preliminary processing and cognition (except video object segmentation as VOS) are invoked only once at the start of each task.
}
  \label{fig:framework}
\end{figure*}

While VLA models can effectively integrate perception and decision-making in an end-to-end manner, they still face significant challenges in real-world applications that require strong generalization capabilities, such as open-vocabulary object manipulation and long-horizon task execution. In contrast to the aforementioned approaches, our work does not primarily focus on improving generalization by directly optimizing the VLA model. Instead, we introduce external modules on top of existing models to form a more comprehensive system, which enhances the performance of the downstream VLA by leveraging external tools for improved information processing.

\subsection{Embodied Agents}
Inspired by the concept of agents~\cite{Luo2025} in the field of large language models, a growing body of research has begun to integrate VLA models as an execution module~\cite{Shi2025, Lei2025, Yang2025, Zhou2025} into embodied agent systems. This is achieved by incorporating additional modules that serve as external tools, effectively expanding the capability boundaries of VLA through appropriate invocation.

The prior work incorporated modules such as task planning, situational memory, and skill libraries. In this paper, we focus on enhancing the agent's tool invocation capability by using web search, object detection, and other functional modules—in combination with current visual observations and task instructions—to identify target objects for manipulation. This approach enables the precise operation of objects beyond the cognitive scope of the single VLA model.

\section{METHOD}

We consist of three major parts, as in Fig.~\ref{fig:framework}: Preliminary Information Processing, responsible for analyzing textual and visual information; Cognition and Memory, responsible for transforming all received information into knowledge accessible to the next part; and Judgment and Execution, responsible for monitoring task progress and interacting with the environment. As shown in the figure, we use LIBERO as the simulation environment.

\subsection{Preliminary Information Processing}

In this part, we employ a planner and a vision pre-processing module to perform the initial decomposition and processing of information.

\subsubsection{Planner} 

The planner is responsible for decomposing complex natural-language instructions into a sequence of structured subtasks executable by downstream modules. To ensure reliability, the planner prompt is designed with strict constraints: each subtask must contain exactly one action verb (e.g., \textit{pick up}, \textit{move}, \textit{open}) and must explicitly specify the relevant objects and locations, with additional syntactic and structural rules enforced so that the post-processing stage can reliably parse the output. This design transforms a complex compound action into multiple smaller subtasks, each consisting of a single action. The planner is implemented using the GLM-4.1V-9B-Thinking \cite{vteam2025glm45vglm41vthinkingversatilemultimodal}, which is locally deployed.
To enable modular extraction of the task list and objects \& locations from GLM’s output, we designed a three-layer post-processing module consisting of: (a) automatic linguistic extraction; (b) error detection and regeneration when extractions fail; and (c) hard-coded task-specific parsing once an error tolerance threshold is exceeded. This architecture ensures that, regardless of what GLM outputs, only valid and high-quality information is passed to the downstream modules.

\subsubsection{Vision Pre-processing} In the initial processing stage of visual information, the framework employs the MMGroundingDINO\cite{zhao2024mmgroundingdino} model to generate a list containing the bounding boxes of the objects and locations provided to this module, as aligned on the first image. Probabilistically, some of the bboxes may be empty due to model failures in recognition or inadequate post-processing. These cases must be further addressed by subsequent cognition and memory.

To better adapt to the overall framework and the task-execution environment, the MMGroundingDINO model is fine-tuned within this framework to improve the accuracy of recognizing task-relevant objects. The experimental setup of this framework is based on the LIBERO simulation environment. Accordingly, 500 randomly rendered images were collected across the LIBERO-Spatial, Goal, Object, and Long datasets \cite{liu2023libero}. Bounding boxes and object names were manually annotated, and data augmentation was applied to the images. Using the MMDetection \cite{mmdetection} toolkit, the model was fine-tuned, resulting in a version that can reliably recognize the objects appearing in these four LIBERO environments.

\begin{figure}[t!]
  \centering
  \includegraphics[width=0.45\textwidth]{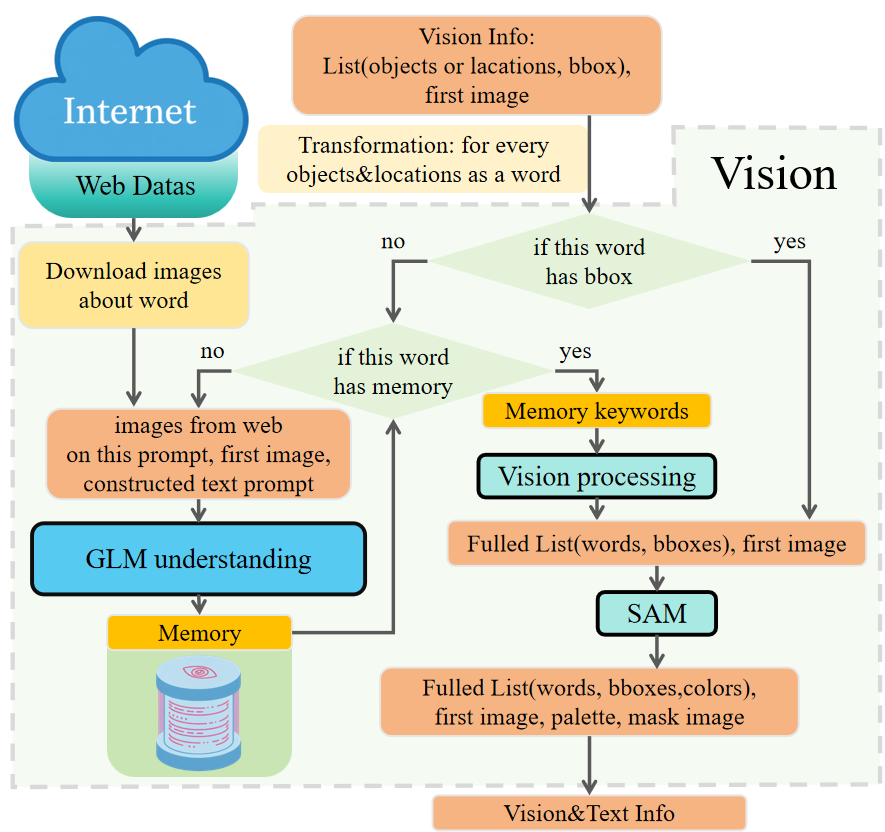} 
  \caption{\textbf{Vision framework.} This figure illustrates the whole structure and contents within \textit{Vision}.
}
  \label{fig:vision}
\end{figure}

\subsection{Cognition \& Memory}

To enhance the out-of-distribution (OOD) performance of the underlying VLA, this project integrates an active web-based information retrieval capability into the higher-level text–image processing pipeline.
The following serial sections will introduce the logic of web search enhancement for visual and linguistic information in detail.

\subsubsection{Vision}

\textbf{Overview.}
In the visual processing pipeline, task–related objects and locations in the third–person robot view are overlaid with transparent, colored masks to reduce reliance on surface textures and mitigate visual overfitting. Fig.~\ref{fig:framework} summarizes this module and its interfaces to the rest of the system. And Fig.~\ref{fig:vision} displayed the detailed logical relationships between the small modules in the vision module.

\textbf{Double judgment.}
For each word (object/location), the system first checks whether a valid bounding box (bbox) is available and, in parallel, whether auxiliary \emph{keywords} are present. If either signal is missing, a visual search branch is triggered: \texttt{bbid}~\cite{ostrolucky2014bulkbingimagedownloader} downloads web images for the word, the images are arranged into a $2\times3$ collage and paired with a structured text prompt, and this input is sent to the GLM Understanding (Vision) module. The resulting keywords, images, and collage are cached in \emph{vision memory} for reuse. The enriched prompt (original text $+$ keywords) is then re-submitted to the detector; if detection still fails, an empty bbox is returned and no mask is applied for that item.

\textbf{GLM understanding (Vision).}
Given the first image, the retrieved web collage, and the current word, this module produces five concise descriptive keywords that anchor the unknown word to elemental attributes (e.g., color, shape, function, size). These keywords support robust re-detection and are stored in memory for subsequent tasks.

\textbf{Vision processing.}
MMGroundingDINO consumes the word together with its keywords to localize the word in the first image, producing a bbox when possible (the “Vision processing” block in Fig.~\ref{fig:vision}).

\textbf{SAM: Segmentation, color encoding, and interface routing.}
Given validated bboxes, SAM2.1-L~\cite{ravi2024sam2} converts each box into a pixel-accurate mask that specifies the target’s location and shape in the image. The outputs (bbox, mask, and the term–color assignment) are packaged with the corresponding \emph{vision memory} (e.g., keywords and web collage). This package is then routed to two consumers: (i) the \textbf{Language} module, which stores the vision-memory fields for the subsequent \emph{replace} step (explained in the next section); and (ii) the \textbf{VOS} pipeline—\emph{a module separate from Vision}—which uses the term–color mapping to guide Cutie~\cite{cheng2023cutie} in generating temporally consistent, color-coded masked image flows. Objects and locations use distinct color palettes so that downstream components can exploit role-aware color cues when learning action–image correspondences.

\textbf{Rationale: instant learning.}
This pipeline converts unfamiliar inputs into familiar representations for MMGroundingDINO, enabling effective OOD generalization by decomposing novel concepts into elemental attributes and anchoring them to known ones. We refer to this as “instant learning”: leveraging prior knowledge to rapidly assimilate unfamiliar concepts. Prior studies indicate that accessible knowledge facilitates the comprehension and memory of new information~\cite{Brod2013}, that successful knowledge construction reactivates previously learned information~\cite{vanKesteren2018}, and that adaptive memory builds on prior knowledge rather than learning tabula rasa~\cite{Bein2020}. Moreover, the explicit color–mask alignment improves visual–text overlap, consistent with findings that finer instance- and token-level alignment boosts performance~\cite{Bi_2023_ICCV} and that stronger color perception benefits color-related generalization~\cite{ColorFoil_2025}.

\begin{figure}[t!]
  \centering
  \includegraphics[width=0.45\textwidth]{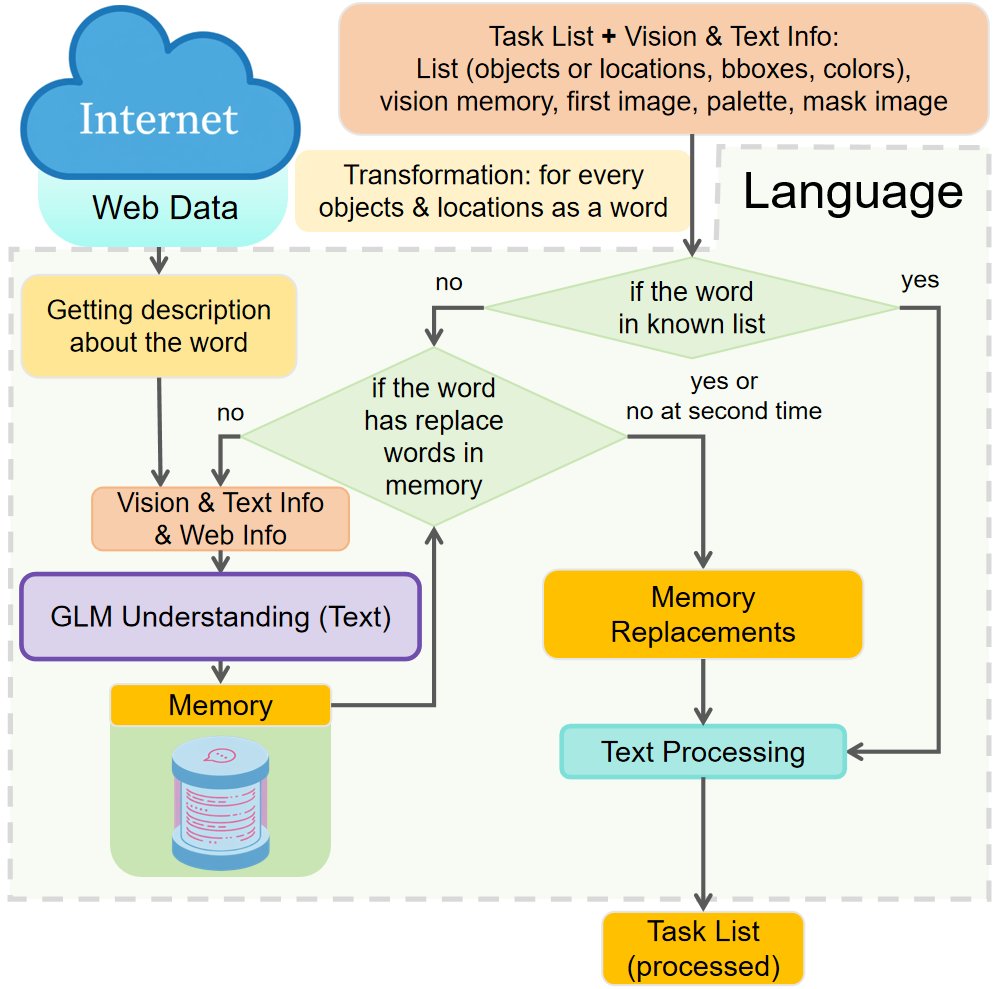} 
  \caption{\textbf{language framework.}  This figure illustrates the whole structure and contents within \textit{Language}.
}
  \label{fig:language}
\end{figure}

\subsubsection{Language}
\textbf{Overview.}
A primary role of the language-processing component is to align all object-related tokens in task prompts with a controlled vocabulary derived from training and fine-tuning, thereby ensuring consistent system-level cognition. The detailed structure and information content of the Language framework are shown in Fig.~\ref{fig:language}.
 
\textbf{Double judgment.} A substitution mechanism handles tokens absent from this vocabulary. For each prompt, once bounding boxes are obtained from the visual pipeline, object terms are replaced at the text level; if no box is detected, substitution is still attempted but designed to return \textsc{None} when no reliable replacement is found. If the token is known on the KnownList (details are at the end of the section~\ref{sub:judgement_and_execution}), it is used directly; otherwise, the GLM (shared with the planner) generates a replacement. 

\textbf{GLM understanding (Text).} The GLM input message comprises: (i) the first image with cropped bounding-box regions and scores (or an empty list), (ii) a collage from web search (or \textsc{None}), (iii) the original prompt, (iv) web-derived keywords (or \textsc{None}), (v) the known-vocabulary list, and (vi) auxiliary descriptive information from external APIs. Analogous to the planner, we designed dedicated input pre-processing and output post-processing modules for the GLM Understanding (Text) component to better align with the language framework and to enable instant learning.

If the replacement word generated by GLM is valid, the corresponding substitution (new corresponding to original) will be recorded in the text memory of the language module, so that when this term reappears for replacement, the system can directly utilize the stored memory. If the replacement word is invalid, no substitution is performed, and no memory is created.

\textbf{Text processing.} Finally, within the current task, once all substitution mappings have been determined, the target terms are replaced accordingly, and the final task list is repaired to eliminate errors arising from long-chain information propagation.

\subsection{Judgment \& Execution}\label{sub:judgement_and_execution}

\textbf{Judgment.} We employ Qwen2.5-VL-3B-Instruct\cite{bai2025qwen25vltechnicalreport} as the verifier. To adapt it more effectively to the experimental scenarios and to improve judgment accuracy, we manually constructed a fine-tuning dataset using videos from the LIBERO dataset. Specifically, video segments were extracted from the original visual recordings of the simulation environment. For each segment, a text prompt was generated corresponding to the current subtask, and annotations were added to indicate whether the subtask had been completed and whether the system could proceed to the next subtask. Fine-tuning of Qwen2.5-VL-3B-Instruct was then carried out using LLaMA-Factory \cite{zheng2024llamafactory} as the training tool, resulting in a verifier better aligned with the LIBERO environments and the task decomposition rules described in the planner section.

Beyond checking whether each subtask is completed, we design a recovery mechanism that uses a dynamic threshold to determine whether the end-effector is stuck or in an anomalous state. Once the recovery detector flags an anomaly, we forcibly set \textit{current task} to ``lift the gripper'' and, after a fixed number of steps, resume the subtask that was active before recovery and restore its execution progress.

\begin{figure*}[t]
  \centering
  \includegraphics[width=0.9\textwidth]{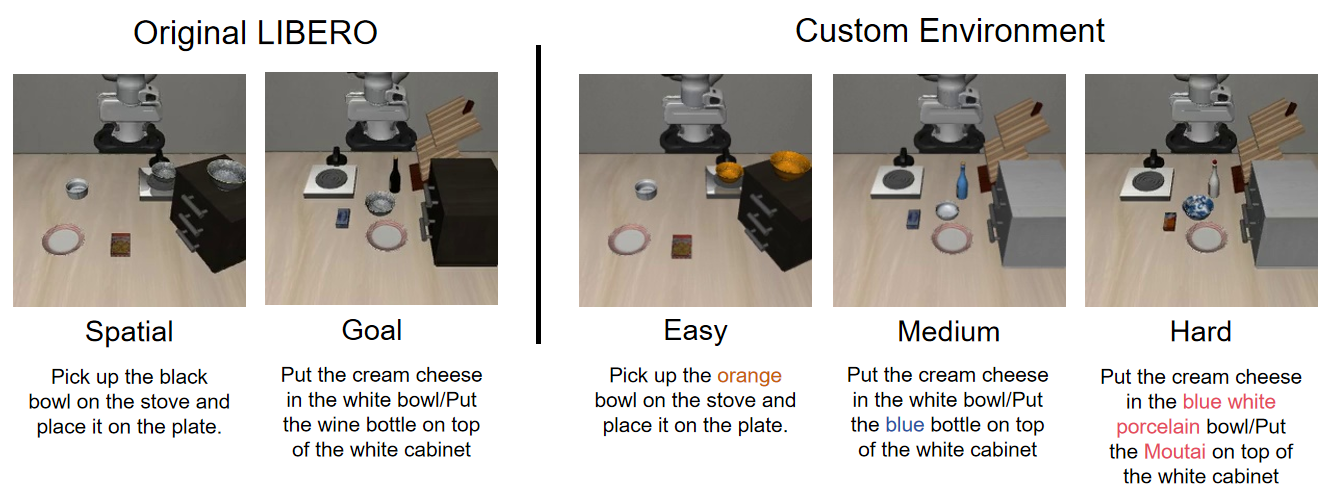}
  \caption{\textbf{Comparison between origin and new environments.} In this figure, we illustrate the differences between the new and original environments. We present a single rendered scene to highlight the modified objects; the novel items appearing in the other scenes share the same appearance.}
  \label{fig:libero-env}
\end{figure*}

\textbf{Execution.} The lower-level VLA is fine-tuned to accommodate the structured inputs produced by the upper-level planner and visual processing modules. In particular, the visual modality of the LIBERO dataset is reformulated by replacing the original third-person RGB videos with RGB videos augmented by transparent colored masks. To construct these masked videos and the accompanying task list, we employ the same vision and language modules described above; all logic and processing remain identical to the main framework. Consequently, during dataset preparation, the vision and language memories already encode the in-distribution(ID) portion of the tasks. For subsequent evaluation on the three OOD environments, any OOD-related memories are re-initialized before each validation run to ensure strict fairness and to isolate the effect of our instant-learning mechanism. Meanwhile, the task descriptions are reformatted into temporally segmented, plan-based task prompts that explicitly reflect the distribution of subtasks over time. Moreover, during fine-tuning and evaluation, the task text prompts are enhanced in the form: ``\textit{now do `current subtask', the whole task is `joint of all subtasks'}'', such that the VLA both knows what it is supposed to do now and what the overall task is. Training the VLA on this modified dataset enables it to process masked visual inputs and sequential subtask prompts consistently with the planner-driven structure, which improves downstream execution performance.

During OpenVLA fine-tuning, a knowledge base of known object terms is built using an NLTK-based extractor. Tokens are identified via tokenization and part-of-speech tagging, aggregated into a JSON vocabulary, and stored with the model for use at inference. This is the \textit{KnownList} in the Language section.

\section{EXPERIMENTS}

We concentrated experiments on evaluating the zero-shot OOD adaptability of the proposed \method~framework. To this end, a new evaluation environment was constructed to specifically test OOD generalization across novel scenarios, in addition to adopting the LIBERO benchmark as a standardized reference. The goal is to examine whether the framework can generalize to previously unseen tasks and maintain robustness without task-specific fine-tuning, while also analyzing the contributions of its key components through ablation studies. Specifically, the experiments aim to answer the following questions: 
\textbf{Q1. }How does the testing performance of \method~on in-domain tasks compare to state-of-the-art VLAs?
\textbf{Q2. }How is the generalization performance of \method~on out-of-distribution test tasks with high difficulty?
\textbf{Q3. }Do the key modules we designed contribute significantly to the framework's generalization performance?

\subsection{Experimental Setup}
\textbf{LIBERO simulation environment.}
Within the original LIBERO simulation environment, we constructed three new variants—Easy, Medium, and Hard—based on the Spatial and Goal environments—comparison between the original and the new environments in Fig.~\ref{fig:libero-env}. The modifications are limited to object appearances as follows. In Easy, the original black bowl was recolored to an \textit{orange series}. In Medium, the black bowl was replaced with LIBERO’s native \textit{white bowl}, the wine bottle was recolored to sky blue and renamed as the \textit{blue bottle}, and the wooden cabinet was replaced with LIBERO’s native \textit{white cabinet}. In Hard, the wine bottle was completely redesigned to resemble the well-known Chinese liquor \emph{Moutai}, the black bowl was redesigned with blue-and-white porcelain patterns and renamed the \textit{blue white porcelain bowl}, and the wooden cabinet was again replaced with the \textit{white cabinet}. The original cream cheese has been replaced with \textit{butter}, which looks different but has approximately the same collision model. No other modifications were introduced beyond these appearance changes.
For the evaluation on the new environments, each task is executed 50 times, and both the overall success rate (SR) and the success rate of each individual task are reported. The same evaluation protocol is applied to the LIBERO original environments when testing the framework.

\textbf{Baseline.}
We compares the proposed \method~framework against several widely recognized, high-performance VLA baselines finetuned on the same LIBERO training dataset: OpenVLA~\cite{kim2024openvla}, OpenVLA-OFT~\cite{kim2025oft}, \(\pi_{0}\)~\cite{black2024pi0}, \(\pi_{0}\text{-FAST}\)~\cite{pertsch2025fast}, and Agentic Robot~\cite{Yang2025}, a embodied agent framework. All experiments are conducted in the original four simulation suites, as well as in the three newly crafted environments specifically designed for OOD evaluation.

\textbf{Training details.}
All components of the framework were trained/fine-tuned on NVIDIA A100--80GB GPUs. For \emph{MMGroundingDINO}, we adopted the default MMDetection training configuration and fine-tuned on our custom dataset using 2 GPUs for 100 episodes. For \emph{Qwen2.5-VL-3B-Instruct}, we used LLaMA-Factory's default \texttt{qwen2-sft} recipe with our custom dataset, increased the number of episodes by a factor of five, and trained on 4 GPUs. For \emph{OpenVLA}, we used the official fine-tuning script on our custom dataset with a learning rate of \(3\times 10^{-4}\), training on 8 GPUs.

\textbf{Implementation.}
This project adopts a 20-step verification waiting period. A custom end-effector jam detection module was implemented with a 10-step recovery waiting to replace the original recovery mechanism and logic. All other model configurations and information transmission pipelines remain the same as described in the \textit{Method} section. In this case, the parameters are closer to those of the original \textit{Agentic Robot}\cite{Yang2025}, making the comparison more meaningful.

\begin{table}
\centering
\caption{\textbf{LIBERO simulation benchmark (Original Environment).} FT denotes fine-tuning on task-specific demonstrations. Bold numbers mark the best \textit{within all classes}. Underline numbers mark the best \textit{within} \textbf{Class 2}.}
\label{tab:sr_original_only}
\setlength{\tabcolsep}{6pt}
\renewcommand{\arraystretch}{1.15}
\begin{tabular}{l cccc c}
\toprule
\textbf{Method} & Spatial & Object & Goal & Long & Average \\
\midrule
\multicolumn{6}{l}{\textbf{Class 1}}\\
\addlinespace[2pt]
OpenVLA-OFT (FT)         & \textbf{97.6} & 98.4 & \textbf{97.9} & \textbf{94.5} & \textbf{97.1} \\
\(\pi_{0}\) (FT)         & 96.8          & \textbf{98.8} & 95.8 & 85.2 & 94.2 \\
\(\pi_{0}\)\text{-FAST} (FT) & 96.4      & 96.8  & 88.6 & 60.2 & 85.5 \\
\cmidrule(lr){1-1}
\multicolumn{6}{l}{\textbf{Class 2}}\\
\addlinespace[2pt]
Agentic Robot            & 85.8 & \underline{89.0} & 81.8 & 61.6 & 79.6 \\
OpenVLA (FT)             & 84.7 & 88.4          & 79.2 & 53.7 & 76.5 \\
\method~(ours)           & \underline{86.4} & 86.2 & \underline{83.2} & \underline{64.4} & \underline{80.1} \\
\bottomrule
\end{tabular}
\end{table}

\subsection{Main Results}

\textbf{Original environments (in-domain; Table~\ref{tab:sr_original_only}).} 
The evaluation shows that Class~1 systems with stronger VLA backbones obtain higher averages. In contrast, our framework uses OpenVLA as the VLA backbone, so the fairest in-distribution comparison is within the OpenVLA family (Class~2). \method~attains the highest Class~2 average SR at 80.1\%, which is higher than Agentic Robot and the fine-tuned OpenVLA. On Object, the SR of our framework (86.2\%) remains below these two baselines. The reason for the result degradation due to a perception bottleneck: 224\(\times\)224 observations and imprecise object names make fine-grained recognition difficult; MMGroundingDINO often misses or mislocalizes boxes; web images used for grounding differ from the simulator views. These perceptual errors can leave the first subtask unresolved, preventing the verifier from advancing and depressing overall SR on affected tasks.

\textbf{Custom environments (out-of-distribution; Tables~\ref{tab:sr_custom_only} and \ref{tab:sr0_transposed}).}
Across the custom environments, all methods exhibit SR declines as OOD difficulty increases, from simple color changes to semantic reinterpretations (e.g., replacing a wine bottle with Moutai) and synonym substitutions (e.g., \emph{plate} $\rightarrow$ \emph{saucer}). Despite this, \method~attains the best overall average SR at 81.5\%. The advantage is most pronounced on the \emph{Hard} environment, where \method~reaches \textbf{76.2}\%, exceeding $\pi_{0}$ by 16.2\% and OpenVLA-OFT by 28.8\% (Table~\ref{tab:sr_custom_only}). Task-level results further highlight robustness on large semantic shifts—for example, \textit{moutai–rack} (72 for \method~vs.\ 44 for $\pi_{0}$) and \textit{bowl–saucer} (88 for \method~vs.\ 16 for $\pi_{0}$), as shown in Table~\ref{tab:sr0_transposed}. These findings support our core premise: by explicitly reforming unfamiliar inputs into the model’s known distribution (via our knowledge-alignment pipeline), \method{} is less perturbed by OOD shifts than competing baselines, even those with more advanced backbones.

\begin{table}[t!]
\centering
\caption{\textbf{LIBERO simulation benchmark (Custom Environment).} SR comparison on Easy/Medium/Hard. FT denotes fine-tuning on task-specific demonstrations. Bold numbers mark the best \emph{across all methods}.}
\label{tab:sr_custom_only}
\setlength{\tabcolsep}{6pt}
\renewcommand{\arraystretch}{1.15}
\begin{tabular}{l ccc c}
\toprule
\textbf{Method} & Easy & Medium & Hard & Average \\
\midrule
\multicolumn{5}{l}{\textbf{Class 1}}\\
\addlinespace[2pt]
OpenVLA-OFT (FT)         & \textbf{98.8} & \textbf{95.4} & 47.4 & 80.5 \\
\(\pi_{0}\) (FT)         & 97.2 & 86.0 & 60.0 & 81.1 \\
\(\pi_{0}\)\text{-FAST} (FT) & 98.0 & 75.2 & 45.8 & 73.0 \\
\cmidrule(lr){1-1}
\multicolumn{5}{l}{\textbf{Class 2}}\\
\addlinespace[2pt]
Agentic Robot (RP)            & 83.8 & 48.6 & 26.2 & 52.9 \\
OpenVLA (FT)             & 85.0 & 66.7 & 32.0 & 61.2 \\
\method~(ours)           & 86.6 & 81.6 & \textbf{76.2} & \textbf{81.5} \\
\bottomrule
\end{tabular}
\end{table}

\subsection{Ablation Study}
We evaluate three ablations in the custom LIBERO-Hard setup, each removing a distinct capability from our framework (Table~\ref{tab:sr0_transposed}). \textbf{w/o mask} excludes the transparent instance/region overlays and \textit{color mask injects}. \textbf{w/o replace} disables lexical normalization, i.e., \emph{unknown or out-of-vocabulary nouns in the task text are no longer substituted} with semantically related in-distribution texts. \textbf{w/o web} turns off all external retrieval and episodic reuse, meaning \emph{no image search, no text search, and no previously cached memory from web retrieval} can be consulted during planning or execution. 
Additionally, we designed an experiment termed \textbf{Agentic Robot (RP)} that removes all the aforementioned modules and replaces every component in the framework~\cite{Yang2025} with the other models mentioned above and additionally omits our subtask-augmentation in the execution prompts, serving as an ablation study.

\begin{table*}[t]
\centering
\caption{\textbf{LIBERO-Hard tasks environment simulation results.} Transposed SR comparison per task. 
The row names under ``\textit{new items}'' (e.g., ``stove'') are concise task abbreviations; ``\textit{new items}'' indicates the number of zero-shot items in the task text prompt. 
\textbf{Bold} marks the best performance across all models. 
\emph{For the Ablation rows, values in parentheses denote the vertical difference from \method~(ours) in the same column, computed as \(\text{Ablation} - \method\).} The Agentic Robot (RP) means w/o mask, replace and web, also no subtask augmentation introduced in the \textbf{Execution} part. Strictly follow the original Agentic Robot pipeline\cite {Yang2025}.
}
\label{tab:sr0_transposed}
\setlength{\tabcolsep}{3pt}
\renewcommand{\arraystretch}{1.2}
\begin{tabular}{l l ccccccccccc}
\toprule
\multirow{3.5}{*}{\textbf{Category}} & \multirow{3.5}{*}{\textbf{\quad Method}} 
  & \multicolumn{1}{c}{\textbf{0 new item}}
  & \multicolumn{5}{c}{\textbf{1 new item}} 
  & \multicolumn{4}{c}{\textbf{2 new items}} 
  & \multirow{3.5}{*}{\textbf{Average} \textbf{SR}} \\
\cmidrule(lr){3-3}\cmidrule(lr){4-8}\cmidrule(lr){9-12}
&
& stove
& \rothead{open-\\drawer} 
& \rothead{drawer-\\bowl} 
& \rothead{saucer-\\stove} 
& \rothead{bowl-\\stove} 
& \rothead{moutai-\\rack} 
& \rothead{bowl-\\saucer} 
& \rothead{bowl-\\cabinet} 
& \rothead{butter-\\bowl} 
& \rothead{moutai-\\cabinet} \\
\midrule
\multirow{3}{*}{\textbf{Class 1}}
& OpenVLA-OFT (FT)   & \textbf{100} & \textbf{100} & \textbf{92} & 8 & 88 & 0 & 0 & 82 & 0 & 4 & 47.4 \\
& \(\pi_{0}\) (FT)   & 98 & 94 & 66 & \textbf{88} & 92 & 44 & 16 & 68 & 0 & 34 & 60.0 \\
& \(\pi_{0}\)\text{-FAST} (FT) & 96 & 62 & 72 & 6 & \textbf{98} & 0 & 34 & \textbf{90} & 2 & 0 & 45.8 \\
\cmidrule(lr){1-2}
\multirow{2}{*}{\textbf{Class 2}}
& OpenVLA (FT)       & 96 & 40 & 14 & 84 & 52 & 0 & 2 & 30 & 2 & 0 & 32.0 \\
& \method~(ours)      & 96 & 78 & 62 & 84 & 86 & \textbf{72} & \textbf{88} & 86 & \textbf{22} & \textbf{88} & \textbf{76.2} \\
\cmidrule(lr){1-2}
\multirow{4}{*}{\textbf{Ablation}}
& \method~(w/o mask)      
  & 94 {\scriptsize(\textcolor{red}{-2})} 
  & 52 {\scriptsize(\textcolor{red}{-26})} 
  & 58 {\scriptsize(\textcolor{red}{-4})} 
  & 78 {\scriptsize(\textcolor{red}{-6})} 
  & 88 {\scriptsize(\textcolor{green}{+2})} 
  & 36 {\scriptsize(\textcolor{red}{-36})} 
  & 84 {\scriptsize(\textcolor{red}{-4})} 
  & 64 {\scriptsize(\textcolor{red}{-22})} 
  & 18 {\scriptsize(\textcolor{red}{-4})} 
  & 76 {\scriptsize(\textcolor{red}{-12})} 
  & 64.8 {\scriptsize(\textcolor{red}{-11.4})} \\
& \method~(w/o replace)   
  & 96 {\scriptsize(0)} 
  & 74 {\scriptsize(\textcolor{red}{-4})} 
  & 26 {\scriptsize(\textcolor{red}{-36})} 
  & 54 {\scriptsize(\textcolor{red}{-30})} 
  & 90 {\scriptsize(\textcolor{green}{+4})} 
  & 16 {\scriptsize(\textcolor{red}{-56})} 
  & 16 {\scriptsize(\textcolor{red}{-72})} 
  & 86 {\scriptsize(0)} 
  & 12 {\scriptsize(\textcolor{red}{-10})} 
  & 42 {\scriptsize(\textcolor{red}{-46})} 
  & 51.2 {\scriptsize(\textcolor{red}{-25.0})} \\
& \method~(w/o web)       
  & 96 {\scriptsize(0)} 
  & 82 {\scriptsize(\textcolor{green}{+4})} 
  & 58 {\scriptsize(\textcolor{red}{-4})} 
  & 82 {\scriptsize(\textcolor{red}{-2})} 
  & 92 {\scriptsize(\textcolor{green}{+6})} 
  & 24 {\scriptsize(\textcolor{red}{-48})} 
  & 84 {\scriptsize(\textcolor{red}{-4})} 
  & 78 {\scriptsize(\textcolor{red}{-8})} 
  & 20 {\scriptsize(\textcolor{red}{-2})} 
  & 36 {\scriptsize(\textcolor{red}{-52})} 
  & 65.2 {\scriptsize(\textcolor{red}{-11.0})} \\
& Agentic Robot (RP)      
  & 96 {\scriptsize(0)} 
  & 38 {\scriptsize(\textcolor{red}{-40})} 
  & 0 {\scriptsize(\textcolor{red}{-62})} 
  & 0 {\scriptsize(\textcolor{red}{-84})} 
  & 44 {\scriptsize(\textcolor{red}{-42})} 
  & 0 {\scriptsize(\textcolor{red}{-72})} 
  & 0 {\scriptsize(\textcolor{red}{-88})} 
  & 64 {\scriptsize(\textcolor{red}{-22})} 
  & 0 {\scriptsize(\textcolor{red}{-22})} 
  & 20 {\scriptsize(\textcolor{red}{-68})} 
  & 26.2 {\scriptsize(\textcolor{red}{-50.0})} \\
\bottomrule
\end{tabular}
\end{table*}

\textbf{Ablation on \emph{mask}.} Disabling transparent masks reduces the average SR from 76.2 to 64.8 (\,$-11.4$\,), with the largest drops on interaction-heavy and cluttered scenes: \textit{open-drawer} $-26$ (78$\to$52), \textit{bowl-cabinet} $-22$ (86$\to$64), \textit{moutai-rack} $-36$ (72$\to$36), and \textit{moutai-cabinet} $-12$ (88$\to$76), see Table~\ref{tab:sr0_transposed}. These patterns indicate the mask overlay is most critical when the VLA must localize within containers/occlusions or discriminate among visually similar instances. Minimal effect on \textit{stove} ($-2$) and even a slight gain on \textit{bowl-stove} ($+2$) suggest that for simple, single-object placements, the raw RGB already suffices, but removing masks consistently hurts spatial reasoning and long-horizon pick-and-place chains.

\textbf{Ablation on \emph{replace}.} Turning off semantic substitution yields the largest average degradation, from 76.2 to 51.2 (\,$-25.0$\,). Catastrophic failures occur when novel or compositional nouns must be grounded: \textit{bowl-saucer} $-72$ (88$\to$16), \textit{moutai-rack} $-56$ (72$\to$16), \textit{moutai-cabinet} $-46$ (88$\to$42), \textit{drawer-bowl} $-36$ (62$\to$26), and \textit{saucer-stove} $-30$ (84$\to$54). These gaps quantify that synonym/alias replacement is the dominant lever for bridging text OOD to the model’s in-distribution vocabulary, especially when two unseen tokens co-occur (the “2 new items” block). Small neutral/positive shifts on \textit{stove} ($0$) and \textit{bowl-stove} ($+4$) imply replacement is unnecessary for well-known nouns, but omitting it severely limits compositional generalization elsewhere.

\textbf{Ablation on \emph{web}.} Removing web image/text search and retrieved memory lowers the average SR to 65.2 (\,$-11.0$\,) and disproportionately harms novel-brand targets: \textit{moutai-rack} $-48$ (72$\to$24) and \textit{moutai-cabinet} $-52$ (88$\to$36). Moderate declines also appear in \textit{bowl-cabinet} $-8$ (86$\to$78). Slight gains on \textit{open-drawer} ($+4$) and \textit{bowl-stove} ($+6$) show that retrieval can inject noise on trivially familiar scenes, but its net benefit on unfamiliar concepts is decisive. Notably, \textit{butter-bowl} remains difficult across settings (ours 22; deltas only $-2$ to $-10$): the low-resolution “butter’’ appears visually ambiguous and cannot be reliably disambiguated by retrieval or text substitution, so even humans struggle to verify it, explaining the uniformly low SR in this task.

\textbf{All three modules removed (\emph{Agentic Robot (RP)}).} This experiment fully adopts the framework~\cite{Yang2025}, with the only modification being the replacement of all corresponding modules with the models used in our proposed method, and also omitting our subtask augmentation, average SR collapses to 26.2 (\,$-50.0$\,). Many hard tasks drop to zero: \textit{drawer-bowl} $-62$ (62$\to$0), \textit{saucer-stove} $-84$ (84$\to$0), \textit{bowl-saucer} $-88$ (88$\to$0), and \textit{butter-bowl} $-22$ (22$\to$0); large losses persist on \textit{moutai-cabinet} $-68$ (88$\to$20), \textit{moutai-rack} $-72$ (72$\to$0), and \textit{open-drawer} $-40$ (78$\to$38). Beyond the ablated capabilities, we find the task-list prompt format used in Agentic Robot introduces substantially increased OOD portion after decomposition (e.g., splitting “put the blue-white porcelain bowl in the cabinet’’ into subgoals that diverge from training distributions). This causes the verifier to repeatedly fail the \emph{first} subtask, preventing progression and yielding SR\,=\,0 for many episodes. In contrast, our prompts condition OpenVLA on “\emph{now do current subtask, while conditioning on the full task context},” which injects stronger ID structure; combined with \emph{mask}, \emph{replace}, and \emph{web}, this design stabilizes execution and underlies the consistent gains in Table~\ref{tab:sr0_transposed}.

\section{CONCLUSIONS}
In this paper, we propose \method, a framework that integrates arbitrary VLAs into a comprehensive embodied agent system. By incorporating modules such as task planning, web search, scene memory, and process verification, our framework enhances the task performance of VLAs. Experiments demonstrate that our module design significantly improves the generalization capability of the agent in grasping objects from unseen concept categories. 

Although our method achieves substantial improvements over existing approaches, it still has certain limitations. Our current framework designs are still confined to relatively rigid procedural structures. Enhancing the versatility of \method~to achieve greater system autonomy and enable the invocation of more external tools to handle a wider range of tasks represents a promising direction for future exploration. Moreover, we have not conducted real-world experiments at this stage, and it is essential to extend our work to open-world real-world grasping evaluations in the future.





\section*{ACKNOWLEDGMENT}
This work was supported by the National Science and Technology Innovation 2030 - Major Project
(Grant No. 2022ZD0208800), and NSFC General Program (Grant No. 62176215).


\bibliographystyle{IEEEtran}
\bibliography{references}

\newpage
\onecolumn 

\section*{\large APPENDIX}

\vspace{1\baselineskip}

\section*{Detailed description of the project framework}\label{app:framework}

\begin{figure}[h]
    \centering
    \includegraphics[width=0.75\linewidth]{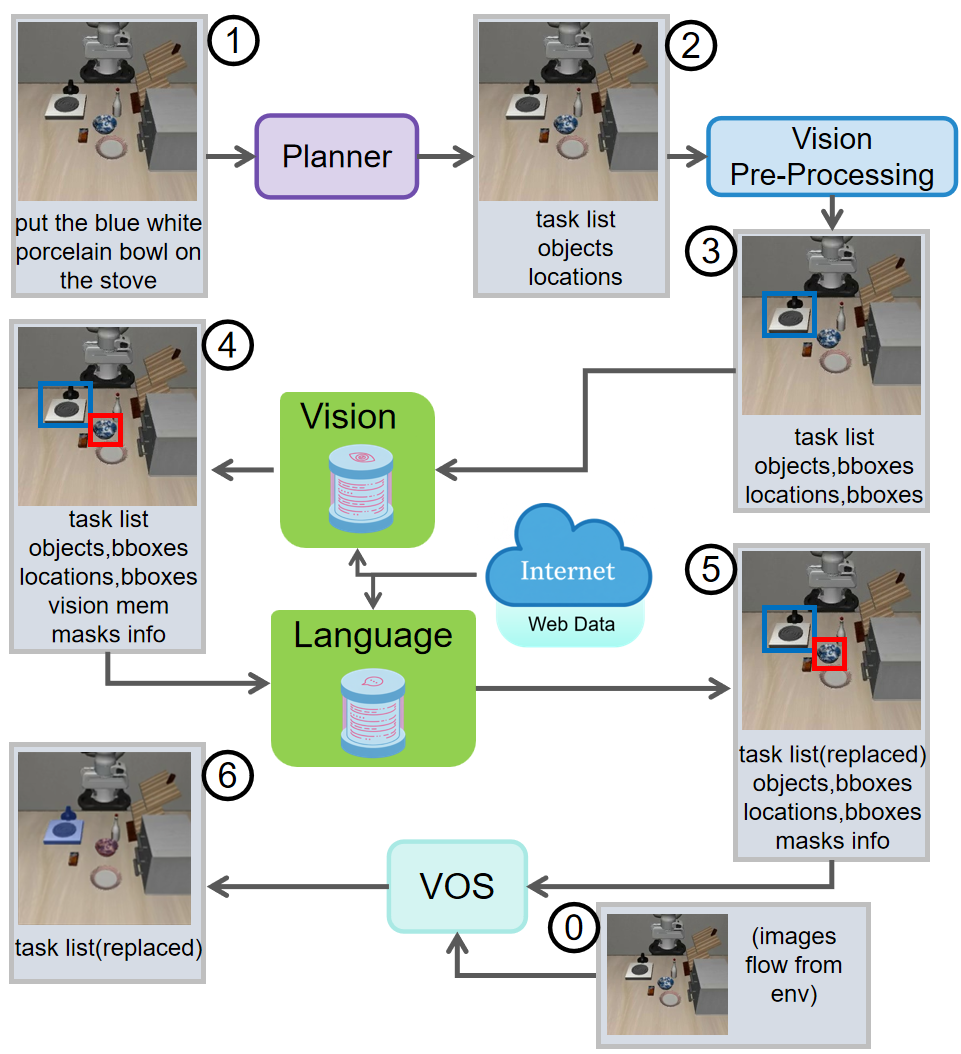}
    \caption{\textbf{Transformation pipeline.} This figure demonstrates how external information is progressively converted into knowledge available to the VLA via the system described in the main text.}
    \label{fig:pipeline}
\end{figure}

In thw Fig~\ref{fig:pipeline} we explain—purely from an information-processing perspective—how all OOD inputs are transformed via the framework described in the main text into ID representations usable by downstream modules; we then outline the design and content of the key prompts used to effect this conversion; finally, we present the computational runtime of each module, so as to evaluate our system’s efficiency.

As illustrated in Fig.~\ref{fig:pipeline}, the environment-sensed information enters the system at \Circled{1}—which matches the typical input format used by VLA systems. Below, we provide a concise, information-processing view of how the content in each gray box transforms and what specific components it comprises.

\begin{itemize}
  \item \Circled{0}\; The environment produces a continually updated image flow. After the task query and the first image are received, only the pathway from \Circled{0} to \Circled{6} remains active for this round; all other transformation pathways are deferred until the next task is initiated.
  \item \Circled{1}\; Here, the image denotes the first frame returned once the environment is ready, and the accompanying text is the task prompt for that environment. In our running example—“put the blue white porcelain bowl on the stove.”—the phrase blue white porcelain bowl denotes a newly introduced object category.
  \item \Circled{2}\; In this information block, the task list is the set of decomposed sub-tasks produced by the planner. For the example in \Circled{1}, the ideal output is: “1) pick up the blue white porcelain bowl; 2) place the blue white porcelain bowl on the stove.” We also extract two structured fields: objects, which are the items that the manipulator must first contact or grasp, and locations, which define the target placement context. In this example, there is one object: “blue white porcelain bowl” and a location: “stove”.
  \item \Circled{3}\; After vision pre-processing, we obtain bounding boxes from a recognition model by using the names in objects and locations together with the image as inputs. This transformation already separates “known” versus “unknown” visual categories: in our example, the stove is known because the model was fine-tuned with stove data, whereas the blue and white porcelain bowl is unknown. This known/unknown status is passed forward to the next Vision module.

\begin{figure}[h!]
    \centering
    \includegraphics[width=0.9\linewidth]{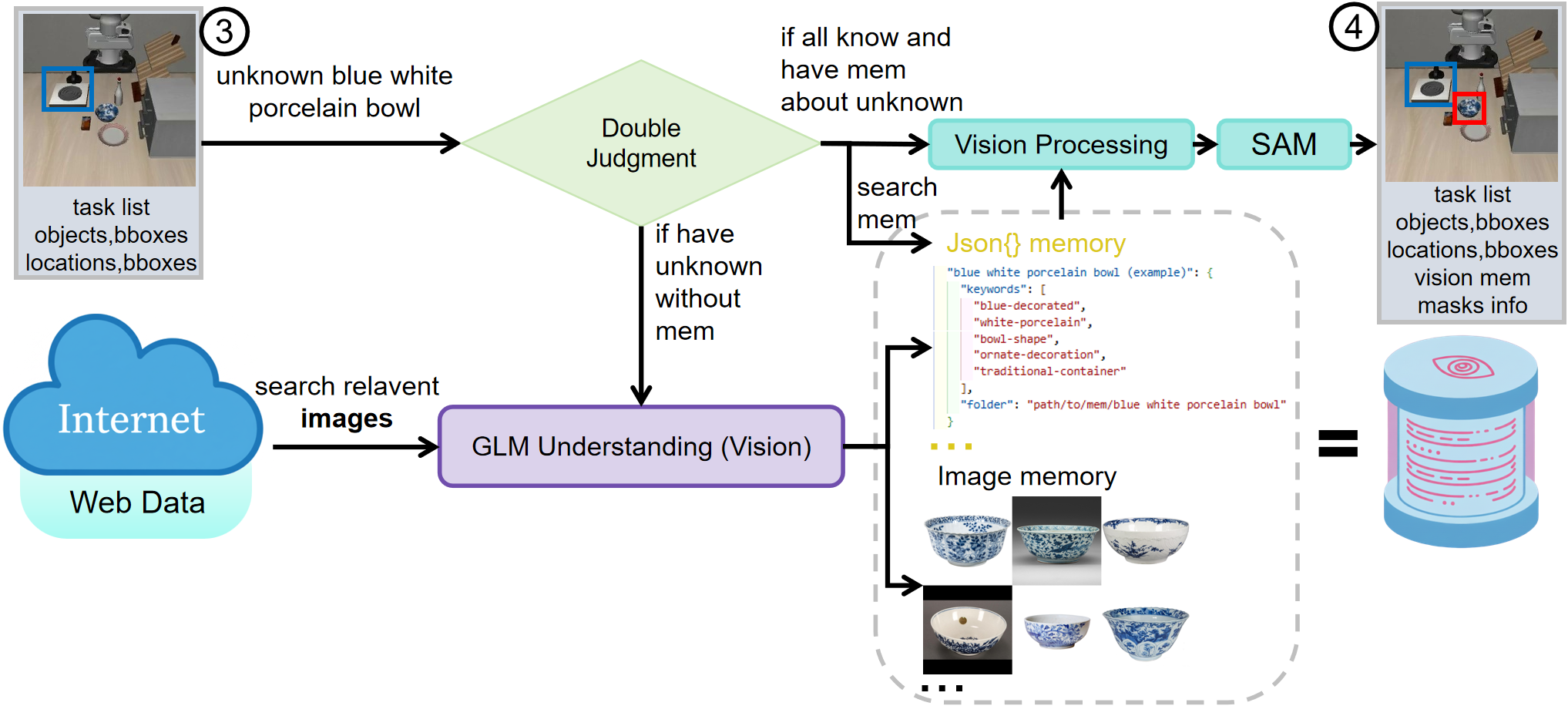}
    \caption{\textbf{Vision processing for unknown blue white porcelain bowl.} The vision memory adopts the same format and similar content as shown in the gray dotted box, for which the equal sign denotes equivalence in structure and content. The system generated the keywords and stored the images here automatically during the evaluation in Table~\ref{tab:sr_custom_only}.}
    \label{fig:Vision_processing}
\end{figure}

  \item \Circled{3}-\Circled{4}\; As shown in Fig.~\ref{fig:Vision_processing}, the information transformation process for the unknown “blue white porcelain bowl” is illustrated. The figure explicates how web‐search images plus the \Circled{3} information are fed into the GLM understanding (Vision) module to generate auxiliary enhanced data for the Vision processing module. In this diagram, we primarily display the storage format of the generated memory and example contents of that memory.

  \item \Circled{4}\; After the Vision stage described in the main text, the module can also recognize some previously unknown categories. In the figure, this is reflected by an additional red bounding box indicating that the blue white porcelain bowl has become identifiable. This recognition is attributed to the cognitive, web-enhanced search phase that creates a persistent memory. Subsequently, all bounding boxes are converted to masks by a SAM-style segmentation step, and masks are color-coded into two palettes corresponding to the objects group and the locations group. The “vision mem” in this block denotes the memory produced by the cognitive search process.

\begin{figure}[t]
    \centering
    \includegraphics[width=0.9\linewidth]{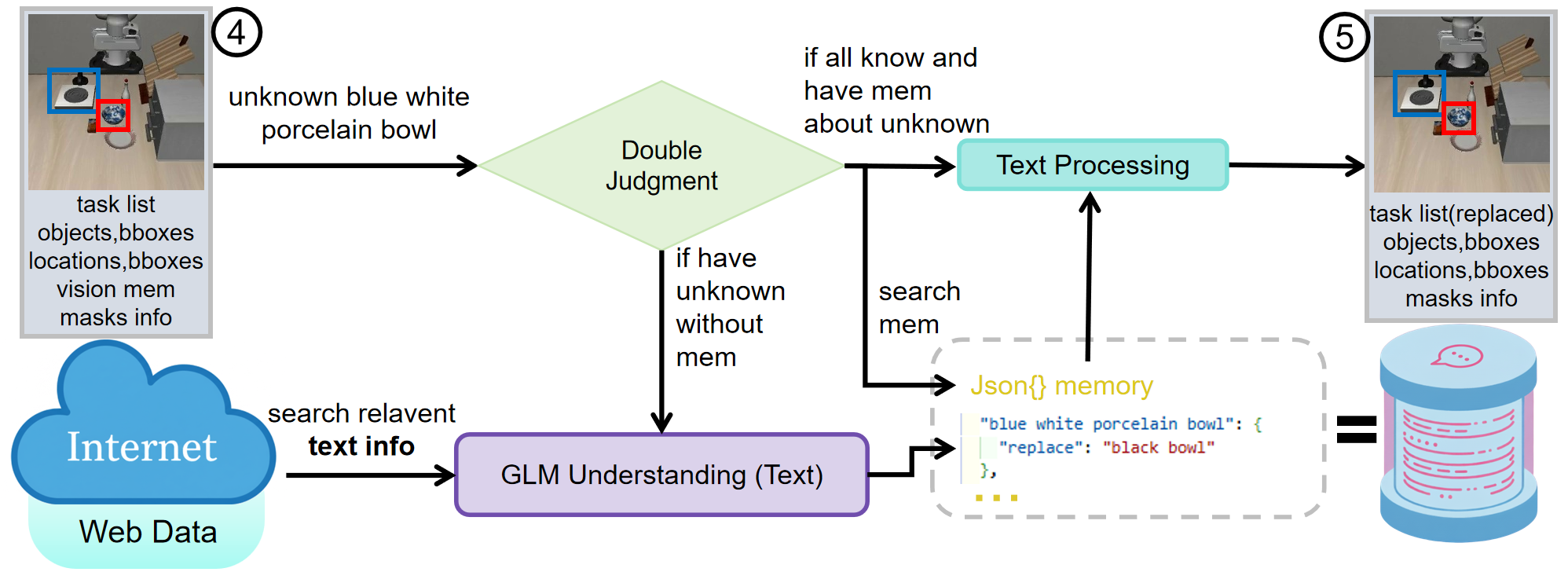}
    \caption{\textbf{Language processing for unknown blue white porcelain bowl.} The equals sign and the gray dotted box denote the same meaning of reference as in Fig.~\ref{fig:Vision_processing}.}
    \label{fig:Language_processing}
\end{figure}

  \item \Circled{4}-\Circled{5}\; As shown in Fig.~\ref{fig:Language_processing}, the overall framework used by the Language module is highly analogous to that of the Vision module. The memory in Language is stored as a JSON file (the “replace map”).

  \item \Circled{5}\; After the Language module, the task list is augmented with color cues and undergoes controlled lexical replacement. In this example, it becomes: “1) pick up the red-mask black bowl; 2) place the red-mask black bowl on the blue-mask stove.” All other metadata remain the same as in \Circled{4}. (Color-aligned masks and text labels are a standard way to synchronize language outputs with pixel-level regions in VLM pipelines.)
  \item \Circled{6}\; This final block serves as the continuously updated input to downstream modules: the image stream is rendered as mask overlays, and the task list is strictly aligned with the visual overlays. Uncertainties at both the language and vision levels are minimized or resolved, yielding a representation that is easier to execute and evaluate. \emph{After task initiation and completion of the cognitive interpretation stage, only the transformation pathway from \Circled{0} to \Circled{6} is retained; the task list is finalized and no longer changes.} In parallel, mask memory distilled from earlier frames is persisted in the VOS, enabling each subsequent frame to infer its masks directly from the new image and the stored mask memory, thereby producing a continuous mask–overlay video stream. 
  Our VOS module is architected following the design principles of the \emph{Cutie} project\footnote{See the Cutie repository for detailed technical specifications and implementation details: \url{https://github.com/hkchengrex/Cutie}.}. For algorithms, data structures, and training/inference pipelines, please refer to that project.
\end{itemize}

Within the Planner, Vision, and Language modules, GLM-4.1V-9B-Thinking is employed. To curb error accumulation from upstream to downstream, we adopt a two-stage failure-handling policy for GLM usage: the first failure triggers an automatic retry, while a second failure invokes a hard-coded fallback or, if necessary, aborts the operation. Consequently, even when truly novel objects cannot be reliably interpreted, the stability of the overall system is preserved.

In every invocation of the GLM and Qwen models, we design prompts tailored to functionality and module interrelations. The planner prompt is shown in \textbf{PLANNER PROMPT}, whose core is the \textit{task\_decomposition\_prompt}, while the other parts enforce module ordering and output constraints. For the verifier, we designed a detailed task-analysis input prompt, as shown in \textbf{VERIFIER PROMPT}. The prompt for GLM understanding (Vision) is given in \textbf{GLM UNDERSTANDING (VISION) PROMPT}. For GLM understanding (Text), as shown in \textbf{GLM UNDERSTANDING (TEXT) PROMPT}, the prompt fed into GLM is not fixed; it is dynamically adapted based on the available inputs and conditions. In all cases, the ultimate objective is to generate a correct replacement mapping from the known vocabulary, given the available context.

\vspace{1\baselineskip}

\section*{Computational efficiency analysis}\label{app:time}

Using the same number of validation runs specified in the Methods (i.e., matching those used to obtain the validation data), we measured and reported the mean computation time per task and per module.

\begin{table}[h!]
\centering
\caption{\textbf{Average computation time.} Computing time in seconds for each module and task.}
\label{tab:time_analysis}
\begin{tabular}{lcccccccc}
\toprule
\textbf{Module} & \textbf{Spatial} & \textbf{Goal} & \textbf{Object} & \textbf{Long} & \textbf{Easy} & \textbf{Medium} & \textbf{Hard} & \textbf{Avg} \\
\midrule
Planner & 20.727 & 19.013 & 17.126 & 25.532 & 21.979 & 19.452 & 20.207 & 20.576 \\
Vision \& Vision Pre-Processing & 0.086 & 0.072 & 0.095 & 0.208 & 0.753 & 1.277 & 1.066 & 0.508 \\
Language & 0.022 & 0.016 & 0.046 & 0.038 & 0.263 & 0.582 & 0.778 & 0.249 \\
VOS & 8.908 & 8.698 & 9.016 & 12.075 & 7.945 & 9.112 & 9.194 & 9.278 \\
VLA & 72.951 & 73.104 & 79.783 & 131.353 & 69.706 & 82.759 & 99.019 & 86.825 \\
Verifier & 2.862 & 3.585 & 3.607 & 5.542 & 4.488 & 4.690 & 4.869 & 4.234 \\
\midrule
\textbf{Total} & \textbf{105.556} & \textbf{104.488} & \textbf{109.673} & \textbf{174.748} & \textbf{105.134} & \textbf{117.872} & \textbf{135.133} & \textbf{121.658} \\
\bottomrule
\end{tabular}
\end{table}

From Table~\ref{tab:time_analysis}, we observe that compared with \cite{Yang2025}, our agentic system’s additional modules—Vision \& vision pre-processing, Language, and VOS—incur only an average extra runtime of \(0.508 + 0.249 + 9.278 = 10.035\) seconds per task over 50 validation runs. This overhead enables the OOD-to-ID conversion pipeline while keeping latency modest. The nearly doubled computation time of the VLA model on the LIBERO-Long tasks arises because every task in that set involves two pick-and-place operations or requires fulfilling two independent actions. Therefore, such tasks demand more steps, resulting in a total runtime roughly twice that of the other three original LIBERO tasks.

\begin{figure}[h!]
    \centering
    \includegraphics[width=0.95\linewidth]{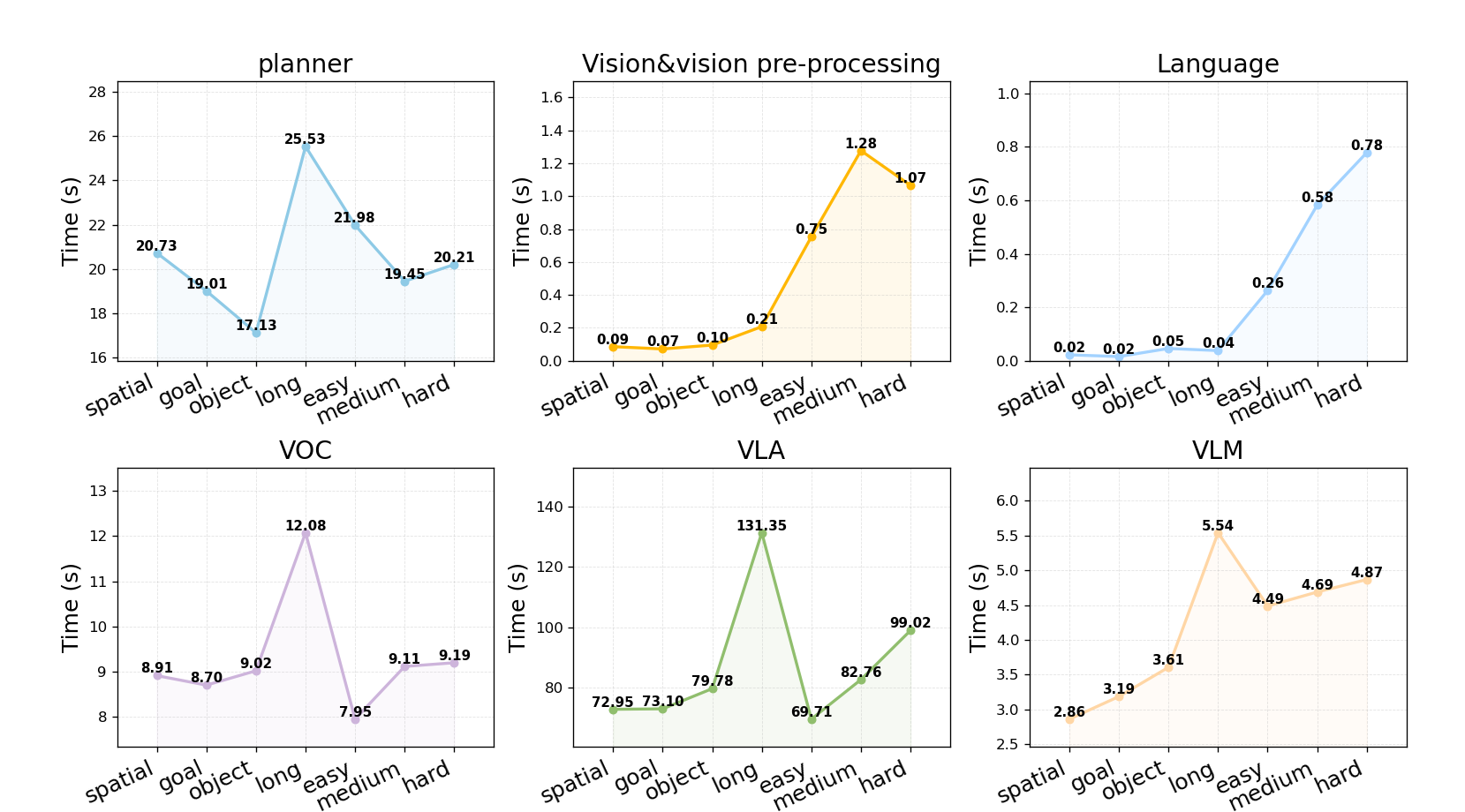}
    \caption{\textbf{Modules runtime across tasks.} This figure shows the average computation time of each module in the agent framework for each task.}
    \label{fig:time_line}
\end{figure}

Because we run GLM-4.1V-9B-Thinking in “thinking” mode, a substantial portion of the Planner’s runtime is spent emitting intermediate “think tokens.” Empirically, we observe that \textit{Planner} latency per task is roughly 20s across different tasks. The Vision and Language modules, which internally embed GLM models, operate under a “first cognition + memory reuse” design: after a correct initial inference, subsequent invocations can reuse stored memory and thus run extremely quickly. As a result, their first-time inference costs are comparable to the Planner (approximately 20s), but repeated usage is much faster. Moreover, in Fig. \ref{fig:time_line}, the modules that execute in every step—VOS, VLA, VLM—show time curves that change in lockstep with task variation, exhibiting nearly identical trend lines. We also note that in our new environment, recognition-centric modules (Vision \& vision pre-processing, Language) incur higher average times due to additional unknown object cognition demands and GLM memory generation. In contrast, Planner—used once per task—shows little runtime difference between the original Libero environment and our custom Libero environment, except for modest variations due to input complexity or error rates.

\section*{LIBERO-Hard task explanation}

In the table~\ref{tab:sr0_transposed} we abbreviate task names; here are their full expansions based on the BDDL filenames in the LIBERO-ZERO environment:

\vspace{1\baselineskip}

\begin{tabular}{ll}
\hline
Abbreviation & Full Human-Readable Task Name \\
\hline
stove & turn on the stove \\
open-drawer & open the middle drawer of the white cabinet \\
drawer-bowl & open the top drawer and put the blue white porcelain bowl inside \\
saucer-stove & push the saucer to the front of the stove \\
bowl-stove & put the blue white porcelain bowl on the stove \\
moutai-rack & put the moutai on the rack \\
bowl-saucer & put the blue white porcelain bowl on the saucer \\
bowl-cabinet & put the blue white porcelain bowl on top of the white cabinet \\
butter-bowl & put the butter in the blue white porcelain bowl \\
moutai-cabinet & put the moutai on top of the white cabinet \\
\hline
\end{tabular}

\vspace{1\baselineskip}
This naming preserves the task structure from the LIBERO-LONG benchmark: each task follows the same schema or template as in the original set, and our version differs only in that we substituted the object terms (e.g., “bowl”, “moutai”) with our custom names.

\newpage

\section*{Planner Prompt}\label{app:Planner Prompt}
\begin{lstlisting}
### reading notice: "#" means the comment in python. This project is written in python, and the following content illustrates the logic and structure of the GLM model prompt. ###

if sign!="success": ### "sign" is a signal for regenerating a better output, sent by the post-processing function. The unsuccessful situations were mainly caused by an unmatchable and unreadable model output. ###
    if sign=="no subtask found":
        additional_info = "PAY MORE ATTENTION TO THE SUBTASKS in your last output, no valid subtask found. You should output the subtask in the same format as the example, without any other analysis or description."
    elif sign=="no objects found":
        additional_info = "PAY MORE ATTENTION TO THE OBJECTS in your last output, no valid objects found in /(here)/. You should output the objects in the same format as the example, without any other analysis or description."
    else:
        additional_info = "PAY MORE ATTENTION TO THE SUBTASKS and OBJECTS in your last output, no valid subtask or objects found. You should output the subtask and objects in the same format as the example, without any other analysis or description."
else:
    additional_info = "You are doing a good job, keep it up"

task_decomposition_prompt =f"""
You are a planning assistant for a fixed robotic arm. Your goal is to break down a high-level task into a sequence of **essential high-level commands**, suitable for a capable Vision-Language-Action (VLA) model to execute directly.

Output Format:
Generate a numbered list of commands. Each command should represent a significant action achieving a clear sub-goal. Stick to the allowed high-level actions.

Example Plan Format (Use **exactly** this level of granularity):
Plan for the robot arm:

Goal: <original instruction>
1. pick up the <object_name_1> /(<object_name_1>)/
2. place the <object_name_1> in the <target_location> /(<object_name_1>,<target_location>)/
3. pick up the <object_name_2> /(<object_name_2>)/
4. place the <object_name_2> in the <target_location> /(<object_name_2>,<target_location>)/

--- Example for a different task ---
Goal: Put the apple in the red bowl
1. pick up the apple /(apple)/
2. place the apple in the red bowl /(apple, red bowl)/

--- Example for another task ---
Goal: Put the cup in the microwave and close it
1. pick up the cup /(cup)/
2. place the cup in the microwave /(cup, microwave)/
3. close the microwave /(microwave)/

--- Example for another task ---
Goal: Turn on the stove and put the pot on it
1. turn on the stove /(stove)/
2. pick up the pot /(pot)/
3. place the pot on the stove /(pot, stove)/

--- Example for another task ---
Goal: Put both books on the bookshelf
1. pick up the red book /(red book)/
2. place the red book on the bookshelf /(red book, bookshelf)/
3. pick up the brown book /(brown book)/
4. place the brown book on the bookshelf /(brown book, bookshelf)/

--- Example for another task ---
Goal: pick the red book near the butter and the brown book on the plate and put them on the left bookshelf
1. pick up the red book near the butter /(red book)/
2. place the red book near the butter on the left bookshelf /(red book, bookshelf)/
3. pick up the brown book on the plate /(brown book)/
4. place the brown book on the plate on the left bookshelf /(brown book, bookshelf)/

--- Example for another task ---
Goal: pick up the yellow and white mug next to the cookie box and place it on the plate
1. pick up the yellow and white mug next to the cookie box /(yellow and white mug)/
2. place the yellow and white mug next to the cookie box on the plate /(yellow and white mug, plate)/

--- Example for another task ---
Goal: put the black bowl in the bottom drawer of the cabinet and close it
1. pick up the black bowl /(black bowl)/
2. place the black bowl in the bottom drawer of the cabinet /(black bowl, cabinet)/
3. close the bottom drawer of the cabinet /(cabinet)/

Instructions:
- Generate **only** high-level commands. 
- Your output should be in the ***ABSOLUTELY SAME format*** as the example above. Even with unseen tasks, follow the same structure. ***WITHOUT ANY OTHER ANALYSIS and DESCRIPTION***.
- **After each command**, include a comment with the object names and locations in */()/*. This is necessary for the VLA model to understand which objects are involved in each command.
- DO NOT include any descriptions of position and order in */()/* (e.g., "first pot", "back of the shelf", "bottom of sth", "upper of sth"), only color and shape are permitted (e.g., "red bowl", "cylindrical box").
    But you should maintain the details of the objects and locations as described in the task to subtask, such as "red bowl near the plate", "brown book on the cabinet", "left bookshelf", "black bowl next to the cookie box", etc.
- **ONLY USE */()/* to EXPRESS *OBJECTS*.** Comments, explanations, and anything else that has nothing to do with expressing objects are not allowed.
- When an object or location has a qualifying modifier, such as a cabinet's drawer, door of a microwave, or the handle of pot, what you are expected to display in the /()/ is actually the **largest specific items these expressions** refer to, which are cabinets, microwaves, and pots, not the parts or subordinate items on these items that belong to these items.
    Meanwhile, you should still maintain the detailed expression in the subtask as "the drawer of the cabinet", "the door of the microwave" (eg. pick up the bottle on the stove; pick up the bowl in the drawer).
- **Allowed commands are strictly limited to:**
    - `pick up [object]`
    - `place [object] on [location]`
    - `place [object] in [location]`
    - `open [object/container/drawer/cabinet/etc.]`
    - `close [object/container/drawer/cabinet/etc.]`
    - `turn on [device]`
    - `turn off [device]`
- Use the commands above **only when necessary** to achieve the goal. Most tasks will primarily use `pick up` and `place`.
- **Explicitly DO NOT include separate steps for:**
    - `locate` (Assume VLA finds the object as part of executing the command)
    - `move to` or `move towards` (Assume the command includes necessary travel)
    - `lift`, `lower`, `grasp`, `release`, `push`, `pull`, `rotate`, `adjust` (Assume high-level commands handle these internally)
- **Assume the VLA model handles all implicit actions:**
    - "pick up [object]" means: Find the object, navigate to it, grasp it securely, and lift it.
    - "place [object] in [location]" means: Transport the object to the location, position it correctly, and release the grasp.
    - "open/close [container]" means: Find the handle/seam, interact with it appropriately (pull, slide, lift) to change the container's state.
    - "turn on/off [device]" means: Find the correct button/switch, interact with it to change the device's power state.
- Use the descriptive names from the task description and **DO NOT make any distortions** in subtasks (e.g., if the task involves {inlist}, make sure the subtasks about them are exactly the same).
- Generate the minimal sequence of these high-level commands required to fulfill the Goal. Ensure the sequence logically achieves the task (e.g., you might need to `open` a drawer before `placing something inside it, even if 'open' isn't explicitly stated in the goal).
- Additional INFO:{additional_info}
Task: {task_description}
Output:
"""
\end{lstlisting}

\vspace{1\baselineskip}
\section*{Verifier Prompt}\label{app:Verifier Prompt}

\begin{lstlisting}

###
The Verifier prompt essentially depends on the input subtask main verb and differentiates each subtask into the following few situations.
###

prefix = (
    f"{title_prefix + ' - ' if title_prefix else ''}"
    f"Observe the inputs (two videos or two image-flow videos). "
    f"The subtask robot arm is currently working on: '{subtask}'. "
)
if verb == "pick up":
    prompt = (
        f"{prefix} Based *Only* on the provided media, has '{object_name}' or anything else been grasped and lifted off any surface by the end? "
        "Answer 'Yes' or 'No'."
    )
elif verb == "place":
    prompt = (
        f"{prefix} Based *Only* on the provided media, has '{object_name}' or anything else been placed '{location_name}' and is the gripper away? "
        "Answer 'Yes' or 'No'."
    )
elif verb in ("turn on", "turn off", "open", "close"):
    target = raw_part or object_name
    action_text = {
        "turn on": "turned on (powered up)",
        "turn off": "turned off (powered down)",
        "open": "fully opened",
        "close": "fully closed",
    }[verb]
    prompt = (
        f"{prefix} Based *Only* on the provided media, has '{target}' or anything else been {action_text} by the end? "
        "Answer 'Yes' or 'No'."
    )
else:
    prompt = (
        f"{prefix} Based *Only* on the provided media, has the instructed action been completed successfully by the end? "
        "Answer 'Yes' or 'No'."
    )
\end{lstlisting}

\vspace{1\baselineskip}
\section*{GLM understanding (Vision) prompt}\label{app:Vision prompt}

\begin{lstlisting}

###
"Query" here means the object or location aiming to be understood.
###

system_prompt = rf"""
    You are an intelligent assistant specialized in analyzing images and extracting meaningful information. Your task is to identify a specific person or object that appears in all provided images and generate five of the most relevant keywords to describe this person or object.
    **Think in ten sentences.** You must follow this rule strictly.
    Guidelines:
    For the combined image:
    If the same person appears in all images:
    Focus on describing the person's gender, skin tone, and occupation.
    Avoid keywords related to clothing or environment.
    Example keywords might include: "female", "light-skinned", "doctor", etc.
    If the same object appears in all images:
    Focus on describing the object's physical characteristics.
    Example keywords might include: "round", "metallic", "small", etc.
    **IMPORTANT** The keywords are going to help another Model to find the same or almost like subjects or persons in the real-world image.
    Thus the keywords should be very specific and descriptive, not general or abstract, and can reflect the basic attributes of this task or thing.
    Making another VLM easily find the same or similar subjects or persons in the real-world image.

    For the current image:
    There is something suitable for the query"{query}", but the model can't find the bbox exactly.
    Your mission is to base on the current image and the combined image to describe the same thing in both.
    
    Output Format:
    Output the keywords in JSON format.
    Ensure the output contains only the keywords, without additional text or explanation.
    The JSON structure should be a list of strings.
    Example JSON Output: ["female", "light-skinned", "doctor", "middle-aged", "smiling"].
    Your output should be in a format that the code below can easily extract the keywords:
    --match = re.search(r"\[.*?\]", output_text[0])
    --  if match:
    --      str_list = json.loads(match.group(0))
    --      print(str_list)

    Task:
    Analyze the provided images and generate five keywords that best describe the identified person or object based on the guidelines above. 
    Output the keywords in the specified JSON format.
    input:{query}
    output:
    """

    messages = [
        {
            "role": "system",
            "content": [
                {
                    "type": "text",
                    "text": system_prompt,
                }
            ]
        },
        {
            "role": "user",
            "content": [
                {
                    "type":"text",
                    "text":"Here is the combined image from the web.",
                },
                {
                    "type": "image",
                    "image": com_image,  ##combined images from internet
                },
            ]
        },
        {
            "role": "user",
            "content": [
                {
                    "type":"text",
                    "text":"This is the current image from the camera.",
                },
                {
                    "type": "image",
                    "image": cur_image,  ##current main view
                },
            ]
        }
    ]
\end{lstlisting}

\vspace{1\baselineskip}
\section*{GLM understanding (Text) prompt}
\begin{lstlisting}
# Build messages for GLM inference (memory-first replace)
messages: list[dict] = []

# 1) System steer (role and objective)
messages.append({
    "role": "system",
    "content": [{
        "type": "text",
        "text": (
            "You normalize open-world object mentions to a closed training vocabulary. "
            "Return EXACTLY ONE label copied verbatim from the allowed list below, "
            "or output NONE if no label applies."
        )
    }]
})

# 2) Allowed vocabulary (verbatim list shown to the model)
allowed_text = "\n".join(f"- {lab}" for lab in known_list)
messages.append({
    "role": "user",
    "content": [{"type": "text", "text": "Allowed vocabulary:\n" + allowed_text}]
})

# 3) The new object mentioned (query term)
messages.append({
    "role": "user",
    "content": [{"type": "text", "text": f"New object mention: {norm_prompt}"}]
})

# 4) Decide available evidence
has_com    = (pil_com is not None)          # composite reference image
has_kw     = bool(keywords)                 # keyword list
has_boxes  = (top_crop is not None)         # highest-score crop from original image
has_scores = bool(boxes_list)               # detector had scores/boxes at all

# 5) Case A: (no comimage, no keywords); include crop if available; else include raw image
if (not has_com) and (not has_kw) and (has_boxes or (pil_image is not None)):
    if has_boxes:
        messages.append({
            "role": "user",
            "content": [
                {"type": "text",  "text": "Evidence crop (highest detector score)."},
                {"type": "image", "image": top_crop},
            ],
        })
    elif pil_image is not None:
        messages.append({
            "role": "user",
            "content": [
                {"type": "text",  "text": "Context image."},
                {"type": "image", "image": pil_image},
            ],
        })

# 6) Case B: (no comimage, no keywords, no boxes/scores); optional raw image only
if (not has_com) and (not has_kw) and (not has_boxes) and (not has_scores):
    if pil_image is not None:
        messages.append({
            "role": "user",
            "content": [
                {"type": "text",  "text": "Context image."},
                {"type": "image", "image": pil_image},
            ],
        })

# 7) Case C: (comimage + keywords + crop are all available); each as its own user turn
if has_com and has_kw and has_boxes:
    messages.append({
        "role": "user",
        "content": [
            {"type": "text",  "text": "Composite reference image from the web."},
            {"type": "image", "image": pil_com},
        ],
    })
    messages.append({
        "role": "user",
        "content": [
            {"type": "text",  "text": "Top-scoring evidence crop from the original image."},
            {"type": "image", "image": top_crop},
        ],
    })
    messages.append({
        "role": "user",
        "content": [{"type": "text", "text": "Image/scene keywords: " + ", ".join(map(str, keywords))}]
    })

# 8) Optional: brief external snippets (web/Wikipedia), one separate turn
if web:
    qs = [norm_prompt] + ([k.strip() for k in keywords] if keywords else [])
    web_brief = fetch_snippets(qs, limit=4)  # function enables searching online and with a "limit" to prevent error content. #
    if web_brief:
        messages.append({
            "role": "user",
            "content": [{"type": "text", "text": "External brief (web/Wikipedia):\n" + web_brief}]
        })

# 9) Final instruction with strict stability constraints
messages.append({
    "role": "user",
    "content": [{
        "type": "text",
        "text": (
            "STRICT CONSTRAINTS:\n"
            "- Output MUST be exactly one label copied verbatim from the allowed vocabulary above, "
            "or the token NONE when no label applies.\n"
            "- DO NOT include any analysis, explanation, reasoning, or additional text.\n"
            "- Format your final decision ONLY as:\n"
            "  <answer>LABEL_OR_NONE</answer>\n"
            "- LABEL_OR_NONE must be one of the allowed labels or NONE."
        )
    }]
})
\end{lstlisting}

\end{document}